\pdfoutput=1

\documentclass[pdflatex,sn-mathphys-ay,iicol]{sn-jnl}%

\usepackage{graphicx}%
\usepackage{multirow}%
\usepackage{amsmath,amssymb,amsfonts}%
\usepackage{amsthm}%
\usepackage{mathrsfs}%
\usepackage[title]{appendix}%
\usepackage{xcolor}%
\usepackage{textcomp}%
\usepackage{manyfoot}%
\usepackage{booktabs}%
\usepackage{algorithm}%
\usepackage{algorithmicx}%
\usepackage{algpseudocode}%
\usepackage{listings}%

\usepackage{multirow}
\usepackage{colortbl}
\usepackage{makecell}
\usepackage[table,xcdraw]{xcolor}
\usepackage{ulem}
\usepackage{hhline}

\theoremstyle{thmstyleone}%
\theoremstyle{thmstyletwo}%

\theoremstyle{thmstylethree}%

\begin{document}

\title[Article Title]{DreamMix: Decoupling Object Attributes for Enhanced Editability in Customized Image Inpainting}

\author[1]{\fnm{Yicheng} \sur{Yang}}\email{2286247133@mail.dlut.edu.cn}
\equalcont{These authors contributed equally to this work.}

\author[1]{\fnm{Pengxiang} \sur{Li}}\email{lipengxiang@mail.dlut.edu.cn}
\equalcont{These authors contributed equally to this work.}

\author*[1]{\fnm{Lu} \sur{Zhang}}\email{zhangluu@dlut.edu.cn}

\author[2]{\fnm{Liqian} \sur{Ma}}\email{LiqianMa.Scholar@outlook.com}
\author[3]{\fnm{Ping} \sur{Hu}}\email{chinahuping@gmail.com}
\author[1]{\fnm{Siyu} \sur{Du}}\email{22409097@mail.dlut.edu.cn}
\author[1]{\fnm{Yunzhi} \sur{Zhuge}}\email{zgyz@dlut.edu.cn}
\author[1]{\fnm{Xu} \sur{Jia}}\email{xjia@dlut.edu.cn}
\author[1]{\fnm{Huchuan} \sur{Lu}}\email{lhchuan@dlut.edu.cn}

\affil*[1]{\orgdiv{Information and Communication Engineering}, \orgname{Dalian University of Technology}, \orgaddress{\city{Dalian}, \postcode{116024}, \country{China}}}

\affil[2]{\orgdiv{Department}, \orgname{ZMO.AI Inc.}, \city{Shenzhen}, \postcode{518000}, \country{China}}

\affil[3]{\orgdiv{Computer Science and Engineering}, \orgname{University of Electronic Science and Technology of China}, \orgaddress{\city{Chengdu}, \postcode{611731}, \country{China}}}




\abstract{Subject-driven image inpainting has recently gained prominence in image editing with the rapid advancement of diffusion models. Beyond image guidance, recent studies have explored incorporating text guidance to achieve identity-preserved yet locally editable object inpainting. However, these methods still suffer from identity overfitting, where original attributes remain entangled with target textual instructions.  To overcome this limitation, we propose DreamMix, a diffusion-based framework adept at inserting target objects into user-specified regions while concurrently enabling arbitrary text-driven attribute modifications. DreamMix introduces three key components: \textit{(i)} an Attribute Decoupling Mechanism (ADM) that synthesizes diverse attribute-augmented image-text pairs to mitigate overfitting; \textit{(ii)} a Textual Attribute Substitution (TAS) module that isolates target attributes via orthogonal decomposition, and \textit{(iii)} a Disentangled Inpainting Framework (DIF) that seperates local generation from global harmonization. Extensive experiments across multiple inpainting backbones demonstrate that DreamMix achieves a superior balance between identity preservation and attribute editability across diverse applications, including object insertion, attribute editing, and small object inpainting.}




\maketitle

\section{Introduction}
\label{sec:intro}
Recent advances in diffusion models~\cite{ddpm, cfg, imagen, dalle_2, sd, sdxl} have catalyzed remarkable progress in image generation, thereby stimulating the evolution in image editing~\cite{ptp, instructpix2pix, lpm, pnp, ijcv_editt1, ijcv_editt2, ijcv_editt3, ijcv_editt4}. Among these applications, subject-driven image customization~\cite{dreambooth, custom_diffusion, ip_adapter, textual_inversion, CharacterFactory} has garnered significant interest, which aims to generate realistic images of a reference object in various contexts. 
To achieve this, existing methods utilize different adaption strategies to enable pre-trained text-to-image diffusion models to apply various editing effects while maintaining the object's identity. 
Despite their efficacy, these methods often focus on regenerating entire scenes, posing challenges when users want to insert specific objects into designated areas within a given scene. This task, known as Subject-driven Image Inpainting, has significant demand in various practical applications, such as effect-image rendering, poster design, virtual try-on, \textit{etc}.  

To address this task, existing methods~\cite{paint_by_example,anydoor,mimicbrush} follow the image inpainting protocol, utilizing a source image and a binary mask as inputs. Additionally, exemplar information is extracted from reference images and used to condition pretrained diffusion models~\cite{sd,sdxl}. However, these reference visual features often encompass overly specific content, thereby constraining the generative capacity of pretrained text-to-image models and complicating the effective editing of objects with desired effects or attributes. Recent efforts~\cite{LAR, tigic} have been made to overcome this limitation by incorporating image and text guidance through attention mechanisms. Nevertheless, challenges remain in balancing training data efficiency with visual editing quality.

\begin{figure}[t]
    \centering
    \includegraphics[width=0.5\textwidth]{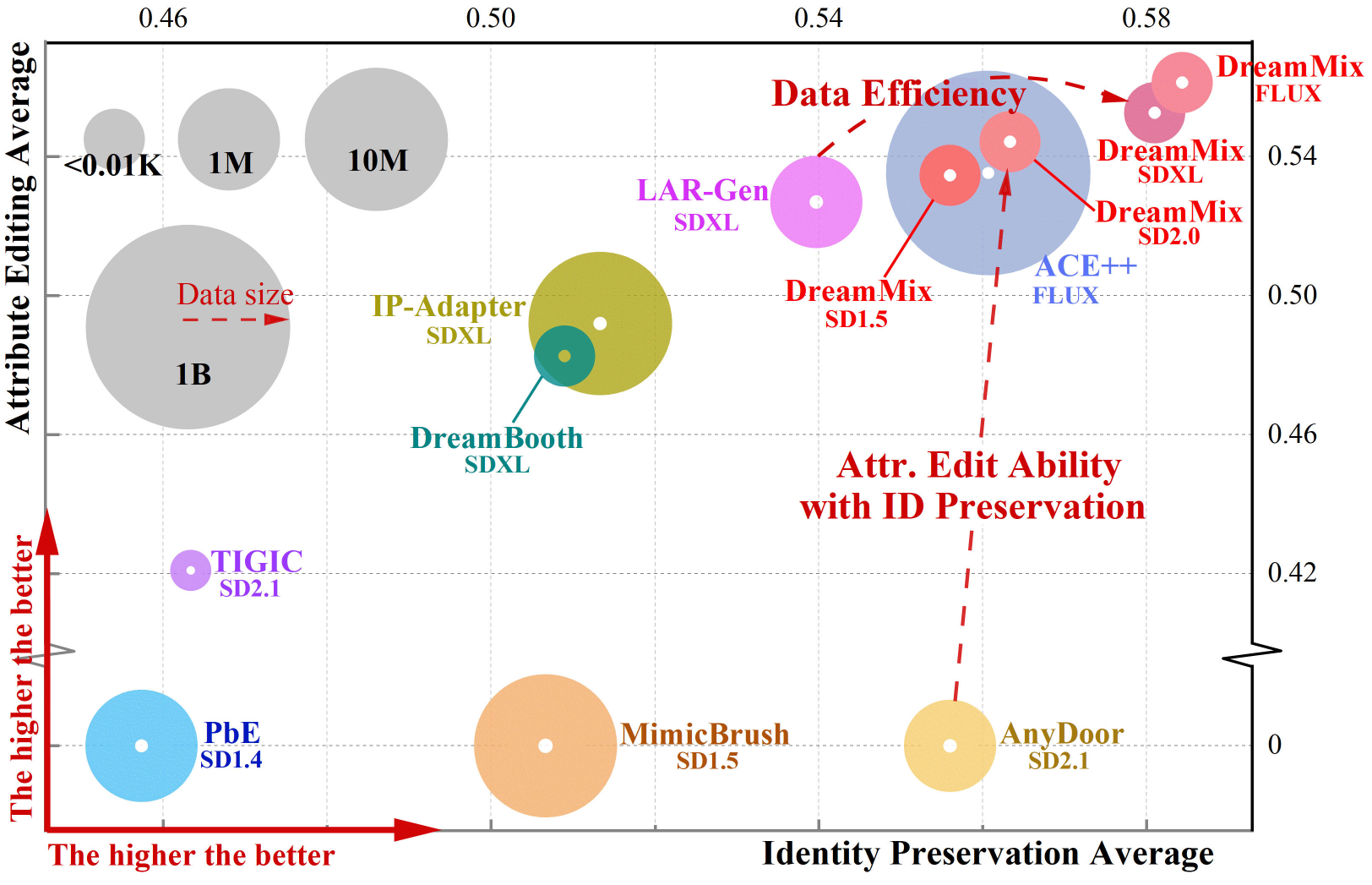}
    \caption{Comparison with state-of-the-art methods in terms of identity preservation, attribute editing, and training data. The reported score is an average on CLIP-T, CLIP-I, and DINO in two applications, respectively. The circle diameter indicates the training data size. Best viewed in color. }
    \label{fig:trade_off_fig}
\end{figure}

In this work, we draw inspiration from recent foundational inpainting models~\cite{sdxl_inpaint, fooocus_inpaint}, which have demonstrated an impressive ability to fill in image regions with high-quality, semantically consistent content based on textual instructions. Leveraging the power of these pretrained text-driven inpainting models for subject-driven inpainting can help alleviate the data-accuracy trade-off, yet is non-trivial. A potential solution is to adopt parameter-efficient image customization techniques (e.g., DreamBooth~\cite{dreambooth}) to adapt models with a few template images. However, this straightforward approach encounters certain limitations in achieving precise object insertion and modification. First, the denoising process in inpainting models may introduce global noise that interferes with local information integration when synthesizing target areas. Second, customization strategies like few-shot fine-tuning~\cite{dreambooth,textual_inversion} usually inject the subject's appearance into a single identity token $[\texttt{sks}]$. However, coupling all object attributes into $[\texttt{sks}]$ may lead to overfitting on the object's appearance, ultimately diminishing the effectiveness of text-based editing instructions.

\begin{figure*}[t]
  \includegraphics[width=\textwidth]{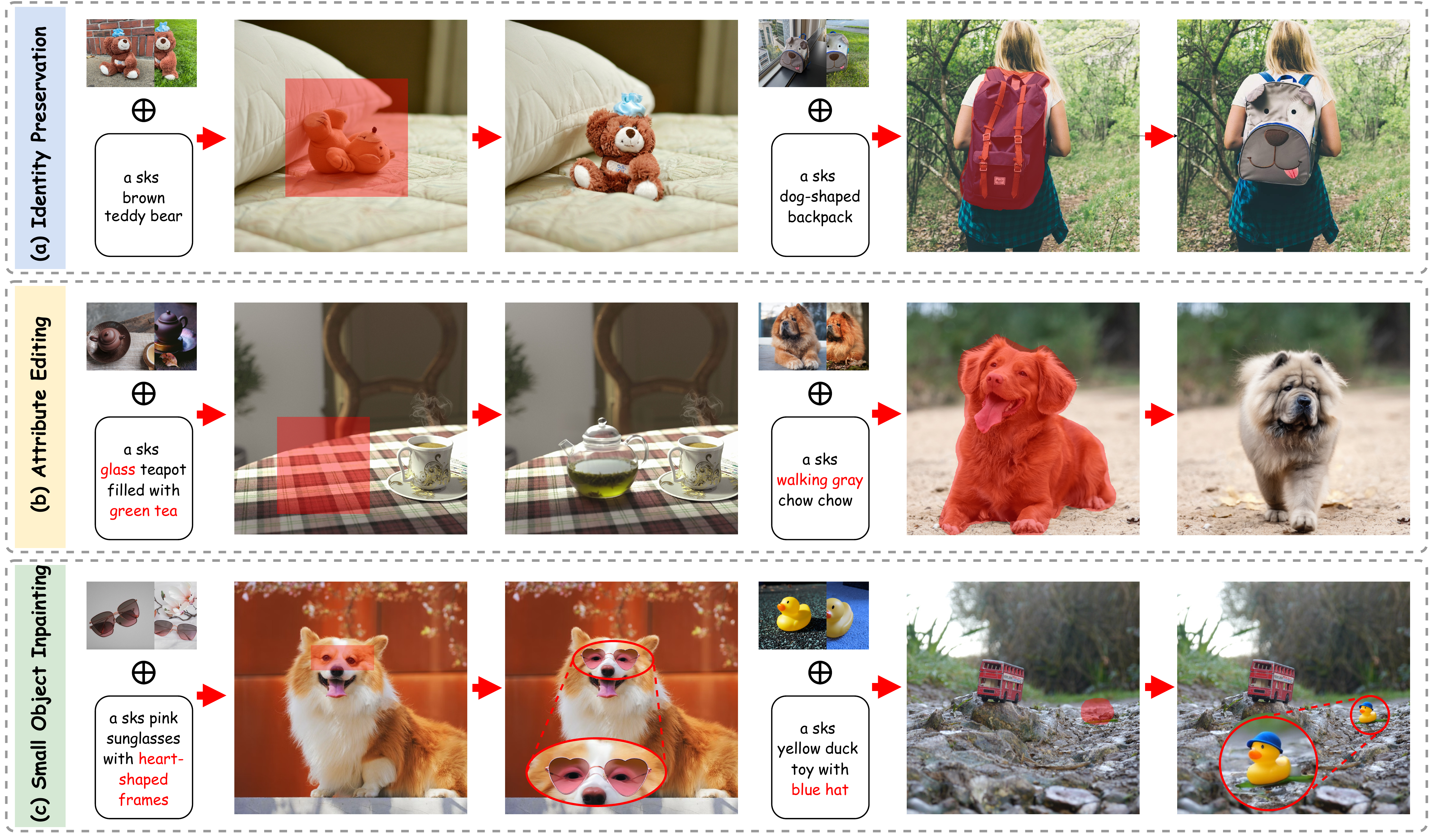}
  \caption{DreamMix on various subject-driven image customization tasks. (a) Identity Preservation: DreamMix precisely inserts a target object into any scene, achieving high-fidelity and harmonized composting results. (b) Attribute Editing: DreamMix allows users to modify object attributes such as color, texture, and shape or add accessories based on textual instructions. (c) Small Object Inpainting: DreamMix effectively performs small object insertion and editing while preserving fine-grained details and visual harmony. }
  \label{fig:teaser}
\end{figure*}

To address these limitations, we propose DreamMix, a diffusion-based generative model for text-subject-driven image inpainting that achieves attribute-level precision and data efficiency. In DreamMix, we introduce two key innovations to resolve overfitting and editing conflicts. First, the Attribute Decoupling Mechanism (ADM) addresses data scarcity and overfitting by leveraging Vision-Language Models (VLMs) to generate attribute-diverse training pairs. By reconstructing prompts and synthesizing corresponding regularization images, ADM disentangles entangled attributes during training, preserving the model’s capacity for flexible editing. Second, the Textual Attribute Substitution (TAS) module enhances inference-time control. Through orthogonal decomposition of text embeddings, TAS isolates and suppresses residual attribute signals from the original object, ensuring that editing instructions dominate the generation process. Additionally, to enhance the subject detail and context harmony in arbitrary inpainting regions, we introduce a Disentangled Inpainting Framework (DIF) on the inpainting backbones, which separates the denoising process into local content generation and global context harmonization.

As shown in Figure~\ref{fig:trade_off_fig} and \ref{fig:teaser}, DreamMix demonstrates superior performance over state-of-the-art methods in both identity preservation and attribute editing, extending editing ability beyond only ID-preserved methods~\cite{mimicbrush,anydoor}, while consuming significantly less training data than large-scale retraining approaches~\cite{LAR,ace_plus}. In addition, the three proposed components are architecture-agnostic, achieving consistent improvement in various inpainting backbones, including SD1.5~\cite{sd15_inpaint}, SD2.0~\cite{sd2_inpaint}, SDXL~\cite{fooocus_inpaint}, and FLUX~\cite{flux_fill}.

Our main contributions can be summarized as follows:
\begin{itemize}
    \item We propose DreamMix, a diffusion-based method that enables precise object insertion and attribute-level editing through synergistic training-inference decoupling of identity and attributes.    
    \item We propose an Attribute Decoupling Mechanism (ADM) to mitigate overfitting by automatically synthesizing attribute-augmented training pairs via VLMs. 
    \item We propose a Textual Attribute Substitution (TAS) module that suppresses source attributes through orthogonal text embedding decomposition, ensuring target instructions dominate generation.
\end{itemize}

\section{Related Work}
\label{sec:related}

\textbf{Image Customization.} 
Recent advances in generative models have significantly improved image customization techniques, allowing for more nuanced and flexible control over the content of generated images~\cite{controlnet, ELITE, instantbooth,ijcv_edit1,ijcv_edit2,ijcv_edit3}.
Recently, subject-driven image customization methods~\cite{textual_inversion, dreambooth, dreamartist, custom_diffusion,CharacterFactory, ominicontrol} have garnered increasing attention
: given a handful of reference images, they adapt a diffusion model to faithfully reproduce a specific subject across diverse scenes and poses with high visual fidelity.
For instance,
Textual Inversion~\cite{textual_inversion} tackles personalization by learning a dedicated text embedding from a small set of images depicting the same subject, providing a lightweight and data-efficient solution but with limited capacity for capturing complex visual details.
DreamBooth~\cite{dreambooth} binds a unique identifier with the target concept and finetunes both the U-Net~\cite{unet} and text encoder~\cite{clip} on subject-specific images, yielding strong identity preservation and effectively preventing overfitting.
OminiControl\cite{ominicontrol} concatenates conditional images and background images, and trains Diffusion Transformer model\cite{flux} based on its self-built large-scale dataset.
However, these methods primarily focus on regenerating entire images but struggle with situations where users want to edit specific local regions of an existing image.

\textbf{Image Inpainting.} Image inpainting\cite{ijcv_inpaint1, ijcv_inpaint2, ijcv_inpaint3} involves reconstructing missing or modified regions of an image. Powered by text-to-image diffusion models, recent works \cite{glide, blend_diffusion, blend_latent_diffusion, powerpaint, smartbrush, sdedit, diffree, repaint, flux_fill} have shown significant improvements in repainting the local regions of existing images with text guidance. For example,
Blend Diffusion \cite{blend_diffusion} combines local text-guided latents with noisy versions of the input image, and a follow-up diffusion step can remap the mix to enhance coherence.
SD-Inpainting \cite{sd} expands the pretrained diffusion models with a concatenation of noise, masked image, and mask as input to produce visually consistent inpainting results.
PowerPaint \cite{powerpaint} enhances the model's ability to support various inpainting tasks by designing learnable task prompts for text-guided object restoration, context-aware image restoration, and object removal, \textit{etc}. 
Flux-fill\cite{flux_fill} uses Diffusion Transformer as the base model and concatenates the background image latent and mask along the channel dimension to achieve high-quality inpainting.
However, despite their effectiveness in text-driven image inpainting, these methods often encounter limitations in compositing user-specific objects into the desired regions of a scene. 

\textbf{Subject-driven Inpainting.} 
Recent methods have started focusing on combining subject-driven image customization with image inpainting to form a task named subject-driven inpainting~\cite{anydoor,tigic,LAR,paint_by_example, mimicbrush}. 
To address this task, some methods~\cite{paint_by_example,anydoor,mimicbrush} utilize an image-guided framework that retrains diffusion models using subject images as templates. However, the absence of original text guidance limits these methods to modify object attributes effectively. This motivates some approaches to develop an image-text-guided architecture that integrates these elements into latent attention layers~\cite{LAR, paint_by_example,tigic}. Nevertheless, most methods~\cite{LAR, paint_by_example} rely on a large-scale training protocol that involves large-scale text-image data construction and a lengthy retraining process. 
While TIGIC~\cite{tigic} is faster, its tuning-free protocol often struggles to provide consistent editing results. 
In this paper, we propose DreamMix, which introduces a disentangled inpainting framework and an attribute decoupling mechanism to achieve data-efficient, editing-enhanced subject-driven image inpainting.

\section{Method}
\label{sec:method}

\begin{figure*}[t]
    \centering
    \includegraphics[width=1\textwidth]{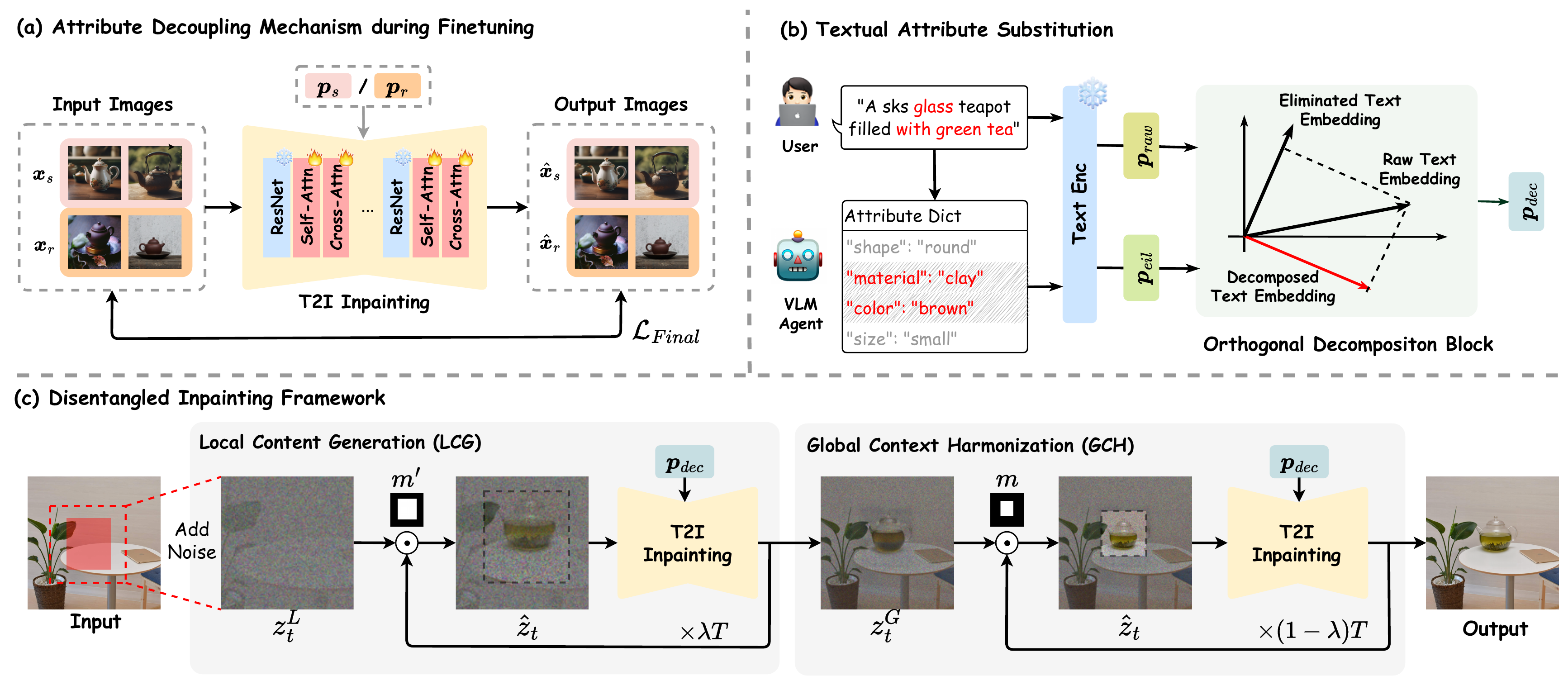}
    \caption{Overview of DreamMix. During finetuning, we use the source data $\{\boldsymbol{x}_s,\boldsymbol{p}_s\}$ along with regular data $\{\boldsymbol{x}_r,\boldsymbol{p}_r\}$ constructed via an attribute decoupling mechanism (Sec.~\ref{sec:prompt_decompose}), to enable pretrained Text-to-Image (T2I) inpainting models to efficiently adapt to subject generation. At testing, we employ a disentangled inpainting framework (Sec.~\ref{sec:DIF}), which divides the denoising process into two stages: Local Content Generation (LCG) and Global Context Harmonization (GCH). Additionally, we propose a textual attribute substitution module (Sec.~\ref{sec:orthogonal_decompose}) to generate a decomposed text embedding to enhance the editability of our method during testing.}
    \label{fig:infer}
\end{figure*}

Given a background image $\boldsymbol{x}$, a binary mask $\boldsymbol{m}$, and a text prompt $\boldsymbol{p}$, DreamMix aims to inpaint the local region of $\boldsymbol{x}$ specified by $\boldsymbol{m}$ to produce an image $\hat{\boldsymbol{x}}$. The refilled areas should not only contain the subject in the reference image(s) $\boldsymbol{x}_s$ but also align with the descriptions provided in the text prompt. 
The overall pipeline of DreamMix is illustrated in Figure~\ref{fig:infer}. It contains three key modules: an attribute decoupling mechanism, a textual attribute substitution module, and a disentangled inpainting framework. We describe each module in detail in the following sections.

\subsection{Preliminaries}
\label{sec:preliminaries}

\textbf{Text-to-Image Diffusion Models.} Diffusion models~\cite{sdxl,sd} generate samples from random noise by learning to estimate the distribution in the reverse process. Specifically, given a random noise $\boldsymbol{\epsilon}\sim\mathcal{N}({0},\mathbf{I})$ and text prompt $\boldsymbol{p}$, text-to-image diffusion models use a noise predictor $\boldsymbol{\epsilon}_\theta$, often U-Net or Transformer model,  to produce a denoised image. Given a dataset of $\mathcal{D}=\{\boldsymbol{x},\boldsymbol{p}\}$, the entire model is trained using the following objective:
\begin{multline}
\mathcal{L_{DM}}\left(\theta;\boldsymbol{x},\boldsymbol{p}\right)=\\
\mathbb{E}_{\boldsymbol{\epsilon}\sim\mathcal{N}({0},\mathbf{I}),t\sim\mathcal{U}({0},\mathbf{T})}\left[||\boldsymbol{\epsilon}_\theta(\boldsymbol{x}_t;\boldsymbol{p},t)-\boldsymbol{\epsilon}||_2^2\right],
\label{eq:train_diffusion}  
\end{multline}
where $t\in [0,T]$ is the time-step.  $\boldsymbol{x}_t=\alpha_t\boldsymbol{x}+\delta_t\boldsymbol{\epsilon}$ where $\alpha_t$ and $\delta_t$ are coefficients that determine the noise schedule of the diffusion process.

\textbf{Personalized Text-to-Image Models.} 
Given a few images of a subject, DreamBooth family~\cite{dreambooth,custom_diffusion} provides a data-efficient protocol for adapting pretrained diffusion models to personalized text-to-image generation. In particular, these methods inject the subject appearance into an identity token [\texttt{sks}] and update the model weights by
\begin{equation}
\mathcal{L}_{\mathcal{DB}}=\mathcal{L_{DM}}(\theta;\boldsymbol{x}_{s},\boldsymbol{p}_{s})+\beta\mathcal{L_{DM}}(\theta;\boldsymbol{x}_{r},\boldsymbol{p}_{r}).
\end{equation}
Here, the first term is the personalization loss for subject appearance modeling, while the second one is the prior-preservation loss, with $\beta$ as a trade-off. $\mathcal{D}_s=\{\boldsymbol{x}_{s},\boldsymbol{p}_{s}\}$ are user-specific subject images and prompts, and $\mathcal{D}_r=\{\boldsymbol{x}_{r},\boldsymbol{p}_{r}\}$ are regularization images and prompts.

\begin{figure*}[t]
    \centering
    \includegraphics[width=0.7\linewidth]{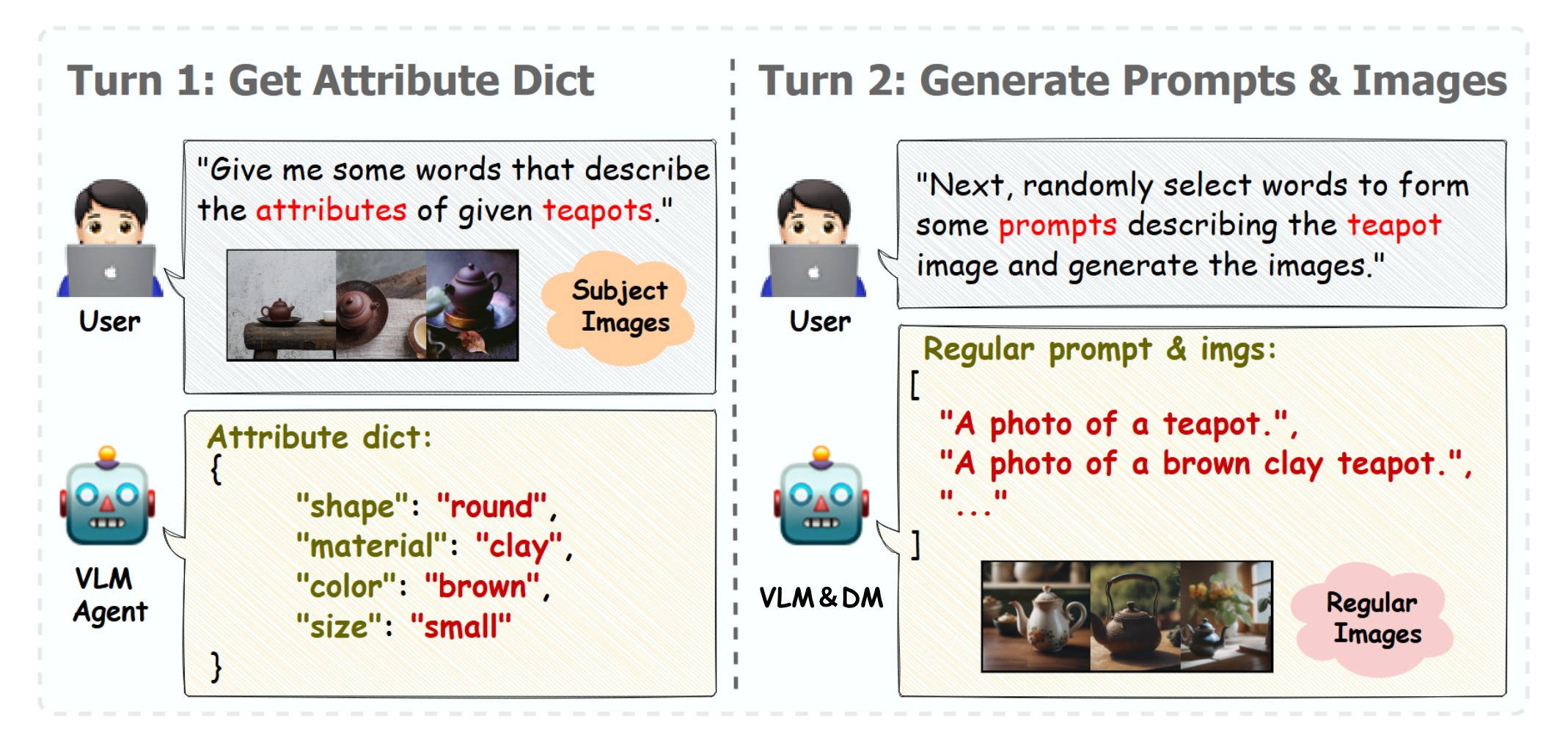}
    \caption{Pipeline of Attribute Decoupling Mechanism (ADM). We obtain the attribute word list using a VLM agent~\cite{chatgpt}, construct regularization data in diverse textual formats, and get regularization image with a diffusion model.}
    \label{fig:fintuning}
\end{figure*}

\subsection{Attribute Decoupling Mechanism}
\label{sec:prompt_decompose}

Few-shot personalization techniques have gained significant attention for their ability to balance data efficiency with generalization capability. However, limitations remain within the few-shot training protocol. On one hand, the textual descriptions used during finetuning are often overly simplified, providing only basic information about the subject's appearance. For example, in DreamBooth~\cite{dreambooth}, an image is typically described as ``a [classname]''. On the other hand, these methods compress all the object's details (e.g., color, texture, and shape) into a single identity token [\texttt{sks}]. 
Under this approach, the features of the subject are confined to the identity token [\texttt{sks}], making it difficult to modify the attributes of the target. Specifically, for a given new attribute, it could be either covered by original features, leaving the subject unchanged, or it may override the original features, causing compromised identity.

In response, we propose an Attribute Decoupling Mechanism (ADM) to address the data scarcity and subject overfitting issues. To circumvent the heavy workload of manual annotation, we leverage advanced Vision-Language Models~\cite{chatgpt} to establish an automatic data reconstruction pipeline. The overall framework of ADM is illustrated in Figure~\ref{fig:fintuning}. Specifically, given a set of subject image-text pairs  $\mathcal{D}_s=\{\boldsymbol{x}_s,\boldsymbol{p}_s\}$, we feed the subject images $\boldsymbol{x}_s$ into a VLM to generate an attribute dictionary that enumerates all attribute words associated with the given subject. Then, we ask the VLM to randomly combine the detected attributes with the original subject to form a series of text prompts with detailed descriptions such as ``a brown clay teapot''. Finally, we generate new images based on the reconstructed text prompts to form a regularization benchmark $\mathcal{D}_r=\{\boldsymbol{x}_r,\boldsymbol{p}_r\}$. Note that different from the subject benchmark $\mathcal{D}_s$, the identity toke \texttt{sks} is removed in the text prompts of regularization benchmark $\mathcal{D}_r$.

Combining source data with regularization data to finetune the pretrained inpainting models can help reduce overfitting, yet is non-trivial. 
DreamBooth finetunes the model parameters beyond just identity tokens, which can lead to attribute words mixing up details.
For example, in ``a brown clay teapot'', the color word might inadvertently bind some texture details, making it difficult to change the texture when transitioning to ``a brown glass teapot'', or influence the precise of texture features when switching to ``a red clay teapot''.
We refer to this phenomenon as ``Concept Infusion''. Similarly, SID \cite{sid} notes that adding detailed descriptions can disrupt the object's identity, aligning with this Concept Infusion issue.

To mitigate this issue, we utilize the regularization data $\mathcal{D}_r$ from ADM for model finetuning and redesign the data reconstruction loss to fit the inpainting task. Specifically, we employ user masks to enforce the inpainting models to fill specific regions. And we propose a loss weight reallocation strategy on Eq.~\ref{eq:loss_re} to reduce the influence of background on feature learning:
\begin{multline}
     \mathcal{L}_{RE} (\theta;\boldsymbol{x},\boldsymbol{p},\boldsymbol{m}) = \tau_1(\boldsymbol{m}\odot||\boldsymbol{\epsilon}_\theta(\boldsymbol{x};\boldsymbol{p})-\boldsymbol{\epsilon}||_2^2)+\\
\tau_2((1-\boldsymbol{m})\odot||\boldsymbol{\epsilon}_\theta(\boldsymbol{x};\boldsymbol{p})-\boldsymbol{\epsilon}||_2^2),
\label{eq:loss_re}  
\end{multline}
where $\tau_1=1.5$ and $\tau_2=0.7$ to emphasize the fill-in regions while suppressing the background area. $\hat{\boldsymbol{x}}=\boldsymbol{\epsilon}_\theta(\boldsymbol{x};\boldsymbol{p})$ represents the pretrained inpainting model that generates the denosied output $\hat{\boldsymbol{x}}$ from a background image $\boldsymbol{x}$, a text prompt $\boldsymbol{p}$, and a binary mask $\boldsymbol{m}$. 
As illustrated in Figure~\ref{fig:infer} (a), we combine both subject data $\mathcal{D}_s$ and regularization data $\mathcal{D}_r$ with this reweight loss to form an overall loss: 
\begin{equation}
  \mathcal{L}_{Final} = \mathcal{L}_{RE} (\theta;\boldsymbol{x}_s,\boldsymbol{p}_s,\boldsymbol{m}_s) + \beta \mathcal{L}_{RE} (\theta;\boldsymbol{x}_r,\boldsymbol{p}_r,\boldsymbol{m}_r),
\end{equation}    
where the first term is for personalized generation and the second is regularization loss. And $\beta=0.4$ is a trade-off.

\subsection{Textual Attribute Substitution }
\label{sec:orthogonal_decompose}

The attribute decoupling mechanism employs advanced VLMs to enhance the diversity of training samples, thereby effectively adapting pretrained inpainting models for subject-driven inpainting. 
However, due to the lack of unseen attribute words during training, relying solely on attribute decoupling mechanism still poses challenges, especially when the target attributes differ significantly from the object identity.
To address this, we introduce a Textual Attribute Substitution (TAS) module during the testing phase to further mitigate the influence of object identity for more precise attribute editing. The pipeline of TAS is illustrated in Figure~\ref{fig:infer} (b).

Specifically, given a text prompt from users, we first query VLMs~\cite{chatgpt} to retrieve the matched attributes from the attribute dictionary that is produced in attribute decoupling mechanism. The selected attribute and the user prompt are then sent to a pretrained text encoder~\cite{clip} to produce their latent embeddings, denoted as $\boldsymbol{p}_{eli}$ and $\boldsymbol{p}_{raw}$. 
Next, we utilize an orthogonal decomposition strategy on the text embeddings to surpass the influence of original attributes in object editing, which is calculated as follows:
\begin{equation}
\boldsymbol{p}_{dec} = \boldsymbol{p}_{raw} - \left(\frac{\boldsymbol{p}_{raw} \cdot \boldsymbol{p}_{eli}}{\|\boldsymbol{p}_{eli}\|}\right) \frac{\boldsymbol{p}_{eli}}{\|\boldsymbol{p}_{eli}\|},
\end{equation}
where \(\boldsymbol{p}_{raw} \cdot \boldsymbol{p}_{eli}\) denotes the dot product of \(\boldsymbol{p}_{raw}\) and \(\boldsymbol{p}_{eli}\), and \(\|\boldsymbol{p}_{eli}\|\) is the norm of \(\boldsymbol{p}_{eli}\). After applying this embedding substitution, the conflicting features of the original object identity are effectively decoupled, making the inpainting model focus on the demand of the target prompt. As shown in Figure~\ref{fig:infer} (c), we apply the decomposed text embedding \(\boldsymbol{p}_{dec}\) to the cross-attention layers at both LCG and GCH stages to boost the editing efficacy of the target text.                   

\subsection{Disentangled Inpainting Framework}
\label{sec:DIF}
In addition to ADM and TAS, we propose a Disentangled Inpainting Framework (DIF) to enhance object insertion in confined local regions. As shown in Figure~\ref{fig:infer} (c), the DIF consists of two seamless stages of Local Content Generation (LCG) and Global Context Harmonization (GCH), to enhance local subject integration and global visual coherence.

\textbf{Local Content Generation.} 
To improve the success of small object insertion, we implement LCG at early time steps of $\lambda{T}$, where $\lambda\in[0,1]$ is a trade-off for time step separation. To improve the generation ability across varying-sized areas, we crop the background image \(\boldsymbol{x}\) into a local patch and obtain the latent code as $\boldsymbol{z}^L=\text{Enc}(\text{Crop}(\boldsymbol{x}_s, \boldsymbol{m}'))$.
Here, \(\text{Crop} (\cdot)\) denotes an image cropping operation guided by a mask \(\boldsymbol{m}'\), which is a slightly enlarged region of original inpainting area, similar to the approach used in AnyDoor~\cite{anydoor}.
$\text{Enc}(\cdot)$ is the VAE encoder. At each time step, the noise predictor $\boldsymbol{\epsilon}_\theta$ takes the text prompt $\boldsymbol{p}_{dec}$ as input to generate the denoised latent code $\boldsymbol{z}_{t}$ by $\boldsymbol{z}_{t}=\boldsymbol{\epsilon}_\theta(\boldsymbol{z}_{t+1};\boldsymbol{p}_{dec},t)$.
During $\lambda{T}$ steps, we exploit a blending strategy~\cite{blend_latent_diffusion} to remain the content of non-editing areas. This process can be formulated as:  
\begin{equation}
\label{eq:LCG_blend}
\hat{\boldsymbol{z}}_{t}=\boldsymbol{z}_{t}\odot \boldsymbol{m}'+ \boldsymbol{z}_t^L\odot (1-\boldsymbol{m}'),
\end{equation}
where $\odot$ denotes element-wise multiplication and $\boldsymbol{z}_t^L$ is the noisy latent code of $\boldsymbol{x}^L$. 
$\hat{\boldsymbol{z}}_{t}$ denotes the blended latent code that will be fed to the noise predictor $\boldsymbol{\epsilon}_\theta$ in the next time step. After the denoising process of $\lambda T$ steps, we project the final result as an RGB image $\hat{\boldsymbol{x}}^L$ and repaste it into the original image $\boldsymbol{x}$ to form an intermediate image $\boldsymbol{x}^G$, which will be used as an image guidance in the global context harmonization stage.

\textbf{Global Context Harmonization.} The LCG in the preceding time steps effectively yields accurate object compositing with a well-defined layout and object appearance. However, without incorporating the global image content, it tends to produce a disharmonious ``copy-paste'' effect in the inpainting areas. To address this issue, we implement GCH on the remaining $(1-\lambda)T$ steps to improve the overall harmony between the generated local region and the background image.
Similar to LCG, we start by transforming the intermediate image $\boldsymbol{x}^G$ into a latent code $\boldsymbol{z}^G$ by $\boldsymbol{z}^G=\text{Enc}(\text{Crop}(\boldsymbol{x}^G, \boldsymbol{m}))$. Here, $\boldsymbol{m}$ is the original user-specific mask. Then, we adopt the same blending strategy in GCH as follows:
\begin{equation}
\label{eq:GCH_blend}
\hat{\boldsymbol{z}}_{t}=\boldsymbol{z}_{t}\odot \boldsymbol{m}+ \boldsymbol{z}_t^G\odot (1-\boldsymbol{m}).
\end{equation}
After executing Eq.~\ref{eq:GCH_blend} in the remaining $(1-\lambda) T$ steps, we obtain the final latent code $\hat{\boldsymbol{z}}_{0}$, which will be fed to VAE decoder to generate the inpainting images.

While DIF uses a blending strategy on explicit input space, its key strength lies in its two-phase optimization mechanism of LCG and GCH. This decoupling prevents the blur-or-conflict trade-off inherent to single-phase blending, ensuring both precise local alignment and seamless global coherence, particularly critical for small-object inpainting.

\begin{figure*}[t]
    \centering
    \includegraphics[width=\textwidth]{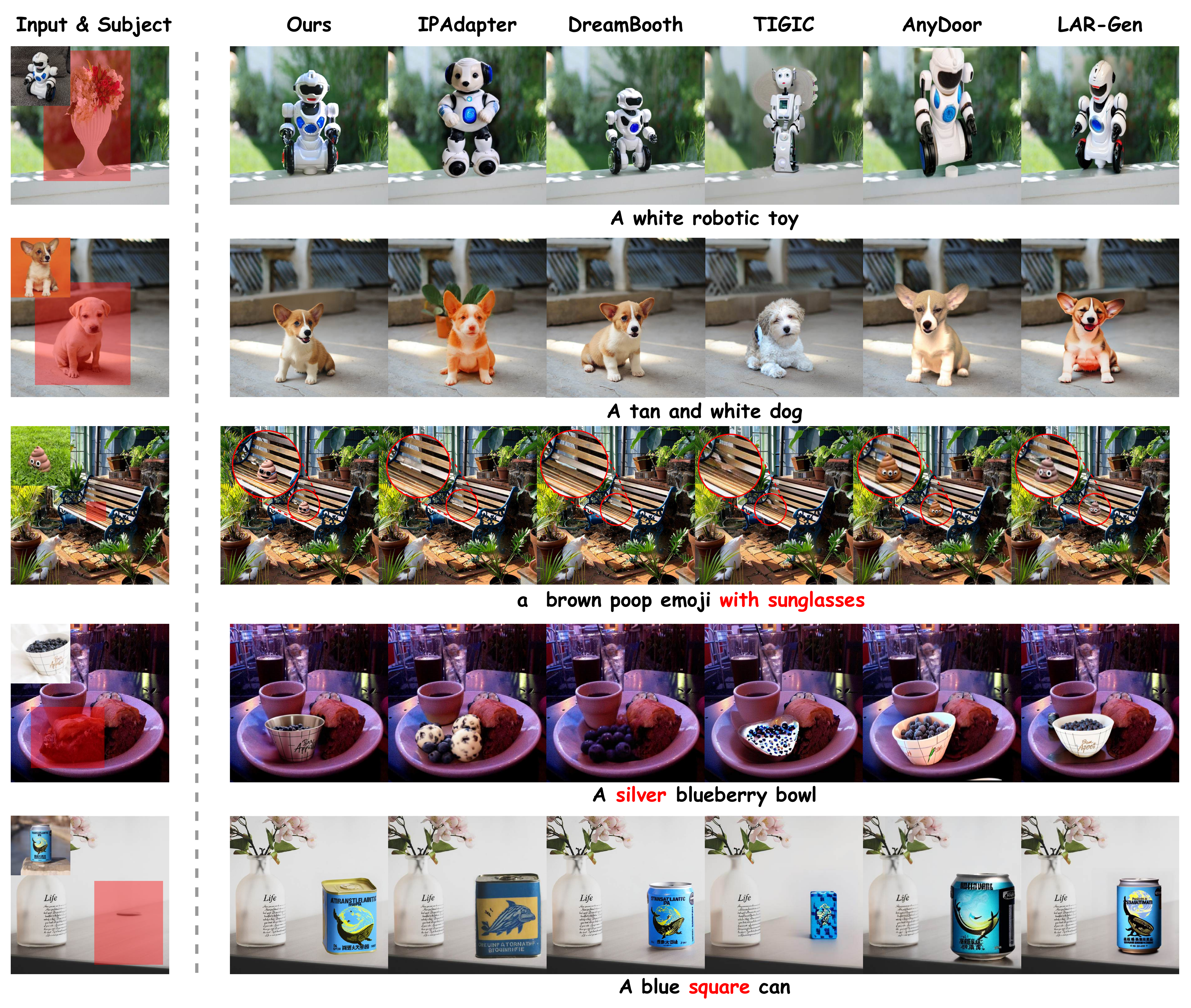}
    \caption{Visual comparison between different methods. From left to right are input background image and subject, visual results of our DreamMix-SDXL, IP-Adapter-SDXL~\cite{ip_adapter}, DreamBooth-SDXL~\cite{dreambooth}, TIGIC~\cite{tigic}, AnyDoor~\cite{anydoor}, and LAR-Gen~\cite{LAR}.}
    \label{fig:compare}
\end{figure*}

\section{Experiments}
\label{sec:experiments}

\subsection{Experiment Setup}
\label{sec:exp_set}

\textbf{Implementation Details.}
{We apply DreamMix on various inpainting backbones, including SD1.5~\cite{sd15_inpaint, sd}, SD2.0~\cite{sd2_inpaint, sd}, SDXL~\cite{fooocus_inpaint}, and FLUX~\cite{flux_fill, flux}}. Note that the SDXL backbone used here is an open-source toolkit developed by~\cite{fooocus_inpaint}, and we re-implement it using the Diffusers~\cite{diffusers} library for fine-tuning.
For all backbones, a LoRA~\cite{lora} with a rank of 4 is applied to \(W_k\) and \(W_v\) in the attention layers. 
The hyperparameter $\lambda$ in DIF is set to 0.7. The VLM used in ADM and TAS is ChatGPT-4o which helps to generate attribute dictionary and formulate 30 text prompts as regularization benchmark.

\begin{figure*}[t]
    \centering
    \includegraphics[width=0.9\textwidth]{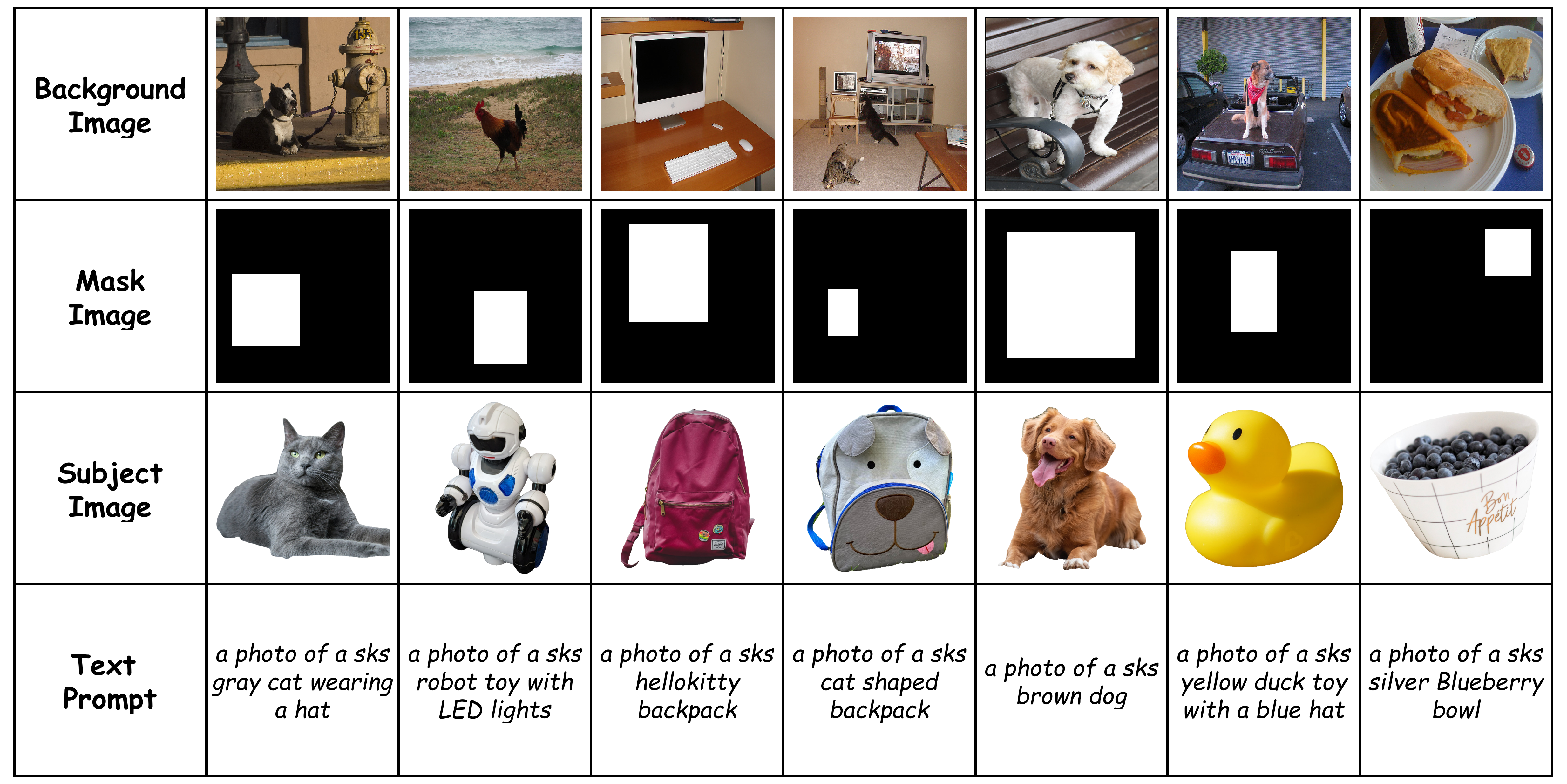}
    \caption{Examples of our benchmark. Row 1: background images to be edited; Row 2: masks marking the inpainting regions; Row 3: objects to be generated in those regions; Row 4: text prompts describing the subjects.}
    \label{fig:benchmark}
\end{figure*}

\textbf{Benchmarks.} We follow prior works~\cite{anydoor,LAR,tigic} to use 30 subjects from DreamBooth~\cite{dreambooth} for performance evaluation. We construct a benchmark that contains over 4,000 high-quality image-mask pairs. Specifically, we collect images from the COCO-val2017 dataset~\cite{coco}, filtering out images or bounding boxes with lower resolutions. We mix the 30 subjects with these background images, resulting in an average of approximately 140 background images per subject. Besides, we manually create 3–5 prompts for identity preservation or attribute editing for each object. We show some examples of our benchmarks in Figure~\ref{fig:benchmark}.

\textbf{Evaluation Metrics.} 
{Since our method employs image and text guidance for simultaneous object composting and text-driven editing, we adopt metrics to evaluate image-text correspondence and image-subject similarity.}
Specifically, to evaluate subject identity similarity, we follow DreamBooth~\cite{dreambooth} to compute CLIP-I \cite{clip} and DINO \cite{dinov2} metrics between generated images and source images. To assess the efficacy of attribute editing, we compute the CLIP-T metric to measure the cosine similarity between generated images and text prompts.
Additionally, we conduct user studies as a subjective evaluation metric.
{To focus on the quality of inpainted regions while considering their integration with the full image, we crop and enlarge the inpainted areas to calculate the evaluation metrics.}

\subsection{Comparison with State-of-the-art Methods}

\begin{table*}[t]
 \caption{Quantitative comparison of different methods on Identity Preservation and Attribute Editing. The ``--" symbol indicates that the method does not support text-driven editing. We label the best and second methods with \textbf{Bold} and \uline{Underlined} styles, respectively.}  
\label{tab:all}
    \centering
    \definecolor{ourmethod}{RGB}{235, 235, 235} 
    \resizebox{\textwidth}{!}{
    \begin{tabular}{l>{\centering\arraybackslash}c>{\centering\arraybackslash}c>{\centering\arraybackslash}c|>{\centering\arraybackslash}ccc|>{\centering\arraybackslash}ccc}
    \toprule
       & & & & \multicolumn{3}{c|}{\textbf{Identity Preservation}} & \multicolumn{3}{c}{\textbf{Attribute Editing}} \\
      \cmidrule(lr){5-7} \cmidrule(lr){8-10}
      \multirow{-2}{*}{\textbf{Method}} & \multirow{-2}{*}{\textbf{Backbone}} & \multirow{-2}{*}{\textbf{Params}} & 
      \multirow{-2}{*}{\textbf{Dataset}} &\textbf{CLIP-T}↑ & \textbf{CLIP-I}↑ & \textbf{DINO}↑ & \textbf{CLIP-T}↑ & \textbf{CLIP-I}↑ & \textbf{DINO}↑  \\ 

    \midrule
    
      PbE~\cite{paint_by_example} & SD1.4 & 1.3B & 1.9M & 0.238 & 0.541 & 0.593 & -- & -- & --  \\
      MimicBrush~\cite{mimicbrush} & SD1.5 & 2.5B & 10M & \uline{0.256} & \uline{0.614} & \uline{0.650} & -- & -- & --  \\ 
      DreamBooth~\cite{dreambooth} & SD1.5 & 1.3B & 4-6 & 0.251 & 0.607 & 0.631  & \uline{0.239} & \uline{0.648} & \uline{0.609} \\ 

      \rowcolor{ourmethod}\textbf{DreamMix} & SD1.5 & 1.3B & 4-6 & \textbf{0.276} & \textbf{0.693} & \textbf{0.699} & \textbf{0.280} & \textbf{0.653} & \textbf{0.671} \\
      
    \midrule

      TIGIC~\cite{tigic} & SD2.1 & 1.3B & 0 & 0.262 & 0.533 & 0.595 & 0.236 & 0.479 & 0.548 \\ 
      
      AnyDoor~\cite{anydoor} & SD2.1 & 2.4B & 0.4M & \uline{0.284} & \uline{0.688} & \uline{0.696} & -- & -- & -- \\ 

       DreamBooth~\cite{dreambooth} & SD2.0 & 1.3B & 4-6 & 0.258 & 0.627 & 0.639 & \uline{0.240} & \uline{0.607} & \uline{0.611} \\ 
      
      \rowcolor{ourmethod}\textbf{DreamMix} & SD2.0 & 1.3B & 4-6 & \textbf{0.284} & \textbf{0.697} & \textbf{0.709}  & \textbf{0.286} & \textbf{0.668} & \textbf{0.679} \\
      
    \midrule

      IP-Adapter~\cite{ip_adapter} & SDXL & 5.6B & 10M & 0.268 & 0.633 & 0.649 & 0.236 & 0.611 & 0.629 \\
      
      LAR-Gen~\cite{LAR} & SDXL & 4.5B & 0.6M & \uline{0.269} & \uline{0.662} & \uline{0.688} & \uline{0.257} & \uline{0.656} & \uline{0.668} \\
      
      DreamBooth~\cite{dreambooth} & SDXL & 3.4B & 4-6 & 0.253 & 0.644 & 0.641  & 0.235 & 0.601 & 0.612 \\ 
      
      \rowcolor{ourmethod}\textbf{DreamMix} & SDXL & 3.4B &  4-6 & \textbf{0.285} & \textbf{0.728} & \textbf{0.730} & \textbf{0.289} & \textbf{0.674} & \textbf{0.695} \\ 
    \midrule

      ACE++~\cite{ace_plus} & FLUX & 16.8B & 0.7B & \uline{0.273} & \uline{0.689} & \uline{0.712} & \uline{0.262} & \uline{0.668} & \uline{0.674} \\ 

      DreamBooth~\cite{dreambooth} & FLUX & 16.8B & 4-6 & 0.261 & 0.670 & 0.691 & 0.253 & 0.655 & 0.664  \\

      \rowcolor{ourmethod}\textbf{DreamMix} & FLUX & 16.8B &  4-6 & 
      \textbf{0.284}  & \textbf{0.741}  & \textbf{0.728} & \textbf{0.290} & \textbf{0.684} & \textbf{0.701} \\ 
      
    \midrule

      Real Image & -- & -- & -- & 0.315 & 0.883 & 0.877 & -- & -- & -- \\
    \bottomrule
    \end{tabular}
    }
\end{table*}

\label{sec:compare_method}
To highlight the efficacy of our method on identity-preserved object compositing and text-driven attribute editing, we conduct comparison experiments on two tasks separately. The quantitative and qualitative comparison results are illustrated in Table~\ref{tab:all} and Figure~\ref{fig:compare}. We select a range of classic and recent methods and categorize them based on the backbone versions. 
For most of the compared methods~\cite{anydoor, mimicbrush,paint_by_example}, we exploit their open-source models to obtain visual results. For LAR-Gen~\cite{LAR}, as the RefineNet is not publicly available, we evaluate it using only the ``Locate'' and ``Assign'' modules on SDXL with an image scale of 1. For TIGIC~\cite{tigic}, we follow their official implementation and use SD-2.1 as the base model. 
For IP-Adapter~\cite{ip_adapter} and DreamBooth~\cite{dreambooth}, we use the same base model as DreamMix to better verify our method's efficacy. For DreamBooth, we employ the same training strategy as used in DreamMix.

\textbf{Identity Preservation.} For identity preservation, we evaluate the ``Real Image'' as an upper bound for all methods, where CLIP-T, CLIP-I and DINO are calculated on the training dataset
As shown in Table~\ref{tab:all}, the proposed DreamMix consistently outperforms other methods on different backbones. 
Anydoor achieves comparable CLIP-T to DreamMix by leveraging large-scale training data and image-guided ID retention to preserve class attributes. However, it struggles with contextual harmonization (lower CLIP-I/DINO) and lacks text-driven editing capabilities, significantly limiting its versatility for dynamic customization tasks. 
%
Additionally,the visual comparison in Figure~\ref{fig:compare} demonstrates that our model outperforms other methods in producing identity-preserved context-harmonious inpainting results.


\textbf{Attribute Editing.} Since some compared methods~\cite{anydoor,mimicbrush,paint_by_example} don't support text-driven editing, we can't report their metrics on this task. As shown in Table~\ref{tab:all}, our method outperforms the other compared approaches on CLIP-I, CLIP-T, and DINO, demonstrating that our model can accurately generate the editing effects of text prompts while preserving the original identity. As shown in the last three rows of Figure~\ref{fig:compare}, our method supports diverse types of attribute editing, such as color, shape, and adding accessories.
{In contrast, other methods struggle with effective subject editing, leading to compromised identity or leaving the target unchanged.}
{Additionally, DreamMix can use various specific masks to enhance fine-grained pose control over editing regions.}
More visual results are provided in the supplementary materials.

\textbf{VLM Metrics in Attribute Editing.} 
We assess attribute-level text–image alignment using the Attribute Binding metric from T2I-CompBench~\cite{t2i_combench}, with ChatGPT-4o serving as the vision–language judge. Specifically, for each edited image, the judge is given the text prompt and the edited output and independently rates three dimensions on a 1–5 scale (higher is better): color, shape, and texture consistency. As summarized in Table~\ref{tab:vlm_merticl_edit}, our method attains strong Attribute Binding scores across all three dimensions, indicating accurate control of color and shape while preserving semantic consistency with the text. 

\begin{table}[t]
\fontsize{7}{8}
\centering
\renewcommand{\arraystretch}{1.1} 
\setlength{\tabcolsep}{4pt} 

\caption{Attribute Binding (T2I-CompBench) scores for attribute editing, with ChatGPT-4o judging color, shape, and texture consistency (1–5; higher is better)}
\label{tab:vlm_merticl_edit}
\begin{tabular*}{\columnwidth}{lc|ccc}
\toprule
\textbf{Method} & \textbf{Backbone} & \textbf{Color↑} & \textbf{Shape↑} & \textbf{Textual↑} \\
\midrule
IP-Adapter    & SDXL    & 1.9  & 2.0 & 2.0\\
LAR-Gen  & SDXL  & 2.4 & 2.7 & 2.5 \\
ACE++  & FLUX  & 2.7 & 2.9 & 2.6 \\
DreamBooth & SDXL        &  2.2 & 2.3 & 2.3  \\
\rowcolor[RGB]{235,235,235} \textbf{DreamMix}    & \textbf{SDXL}   & \textbf{3.3} & \textbf{3.6} & \textbf{3.2} \\
\bottomrule
\end{tabular*}
\end{table}

\textbf{User Study.} 
We conduct a user study to further evaluate the feasibility of our method on identity preservation and attribute editing. We ask 100 volunteers to complete a questionnaire consisting of 30 questions, resulting in a total of 3000 responses. In each question, we present an image of a reference object, a background image, a target text, and the images generated by anonymous methods. {Participants are asked to answer the following questions: ``Which of the following images is the best by considering Image Quality (IQ), Image Similarity (IS) and Text Relevance (TR)?''}
We summarize the results and present them in Table~\ref{tab:user_study}. As shown, the proposed DreamMix has an overwhelming preference for both identity preservation and attribute editing.

\begin{table}[t]
    \centering
\caption{User study results for \textit{Identity Preservation} and \textit{Attribute Editing} tasks, showing user preference percentages for each method. DreamMix outperforms other methods significantly in both categories, indicating strong user preference.}
    \label{tab:user_study}
    
    \fontsize{7}{8}

    \renewcommand{\arraystretch}{1.1} 
    \setlength{\tabcolsep}{1pt} 

    \begin{tabular*}{\columnwidth}{l>{\centering\arraybackslash}c| >{\centering\arraybackslash}c>{\centering\arraybackslash}c| >{\centering\arraybackslash}c >{\centering\arraybackslash}c >{\centering\arraybackslash}c}
    
    \toprule
     & & \multicolumn{2}{c|}{\textbf{ID Pres.}} & \multicolumn{3}{c}{\textbf{Attr. Editing}} \\
    \cmidrule(lr){3-4} \cmidrule(lr){5-7}
    \multirow{-2}{*}{\textbf{Method}} & \multirow{-2}{*}{\textbf{Backbone}} & \makebox[0.05\textwidth]{\textbf{IS} ↑} & \makebox[0.05\textwidth]{\textbf{IQ} ↑} & 
      \makebox[0.05\textwidth]{\textbf{TR} ↑} & \makebox[0.05\textwidth]{\textbf{IS} ↑} & \makebox[0.05\textwidth]{\textbf{IQ} ↑} \\ 
    \midrule
    
    TIGIC  & SD2.1 & 3\% & 2\% & 1\% & 2\% & 1\% \\
    LAR-Gen & SDXL & 19\% & 21\%  & 13\% & 12\% & 13\% \\
    IP-Adapter & SDXL & 4\% & 8\% & 5\% & 6\% & 8\% \\
    DreamBooth & SDXL & 14\% & 15\% & 10\% & 7\% & 11\% \\
    \rowcolor[RGB]{235, 235, 235} \textbf{DreamMix} &  \textbf{SDXL}  &  \textbf{60\%}  &  \textbf{54\%} &  \textbf{71\%} & \textbf{73\%} & \textbf{67\%} \\
    \bottomrule
    \end{tabular*}
\end{table}

\subsection{Ablation Studies}
\label{sec:ablation_study}

\begin{figure*}[th]
    \centering
    \includegraphics[width=0.6\textwidth]{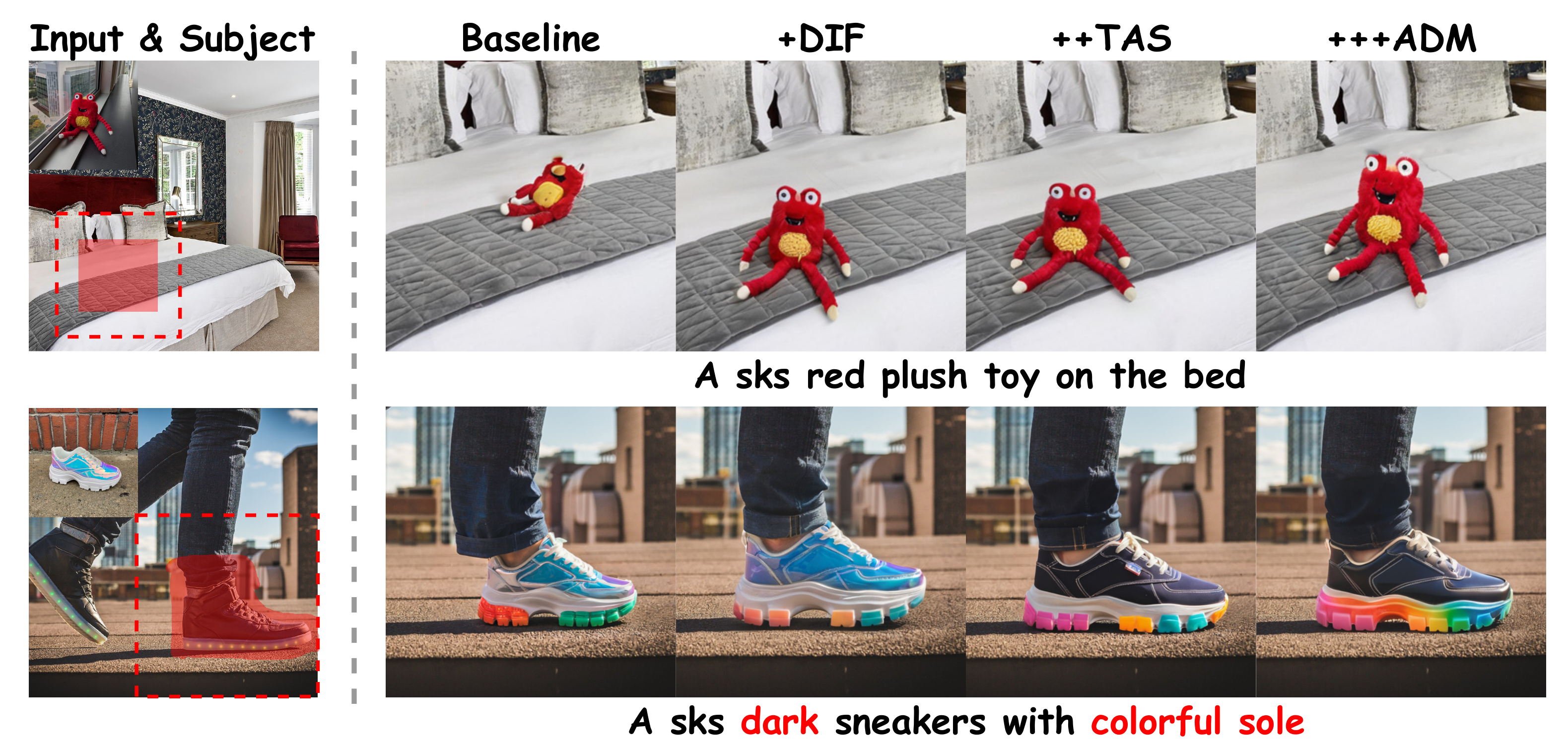}
    \caption{Visual examples for ablation studies on identity preservation (top row) and attribute editing (bottom row).}
    \label{fig:ablation}
\end{figure*}

\textbf{Effectiveness of Each Component.} 
We conduct experiments to verify the effectiveness of Textual Attribute Substitution (TAS) module, Attribute Decoupling Mechanism (ADM) and Disentangled Inpainting Framework (DIF) in DreamMix. The quantitative results are presented in Table~\ref{tab:ablation}. 
We use DreamBooth~\cite{dreambooth} on our SDXL inpainting model~\cite{fooocus_inpaint} as a baseline method.
We incrementally incorporate each component and evaluate them on both tasks. The results in Table~\ref{tab:ablation} demonstrate that the proposed DIF effectively improves the generative capacity of the baseline method in both tasks. 
For attribute editing, our model achieves a continual increase in CLIP-T as the incremental involve of TAS and ADM, while exhibiting a slight decrease in CLIP-I and DINO. These results indicate that the edited regions are more consistent with the text prompts and, consequently, slightly differ from the original appearance in the feature space.
This demonstrates the effectiveness of the proposed TAS and ADM in improving text-driven editing. 
In addition, the results on identity preservation demonstrate that incorporating TAS and ADM does not affect identity preservation if no editing prompts are given. 
Figure~\ref{fig:ablation} illustrates the visual examples of ablation experiments. In the first row, we observe that DIF enhances image quality, while TAS and ADM do not negatively impact the subject’s identity. In the second row, DIF establishes a solid foundation that allows TAS and ADM to effectively improve editing precision.   

\begin{figure*}[th]
    \centering
    \includegraphics[width=0.6\textwidth]{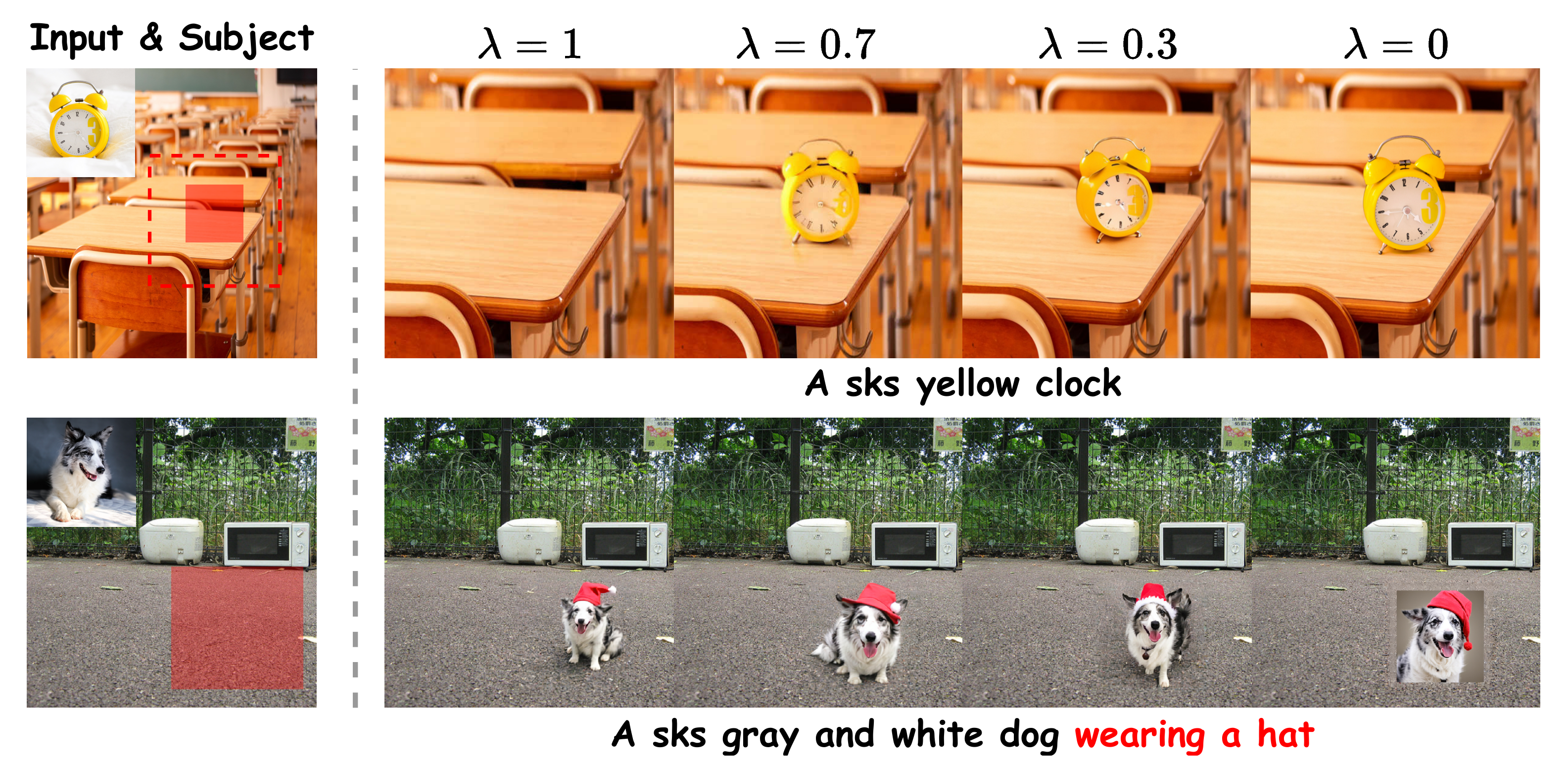}
    \caption{Effect of different values of \(\lambda\) in disentangled inpainting framework. \(\lambda=1\) means only GCH stage is performed while  \(\lambda=0\) means only LCG stage is used. $\lambda$ is set to 0.7 in our experiments.}
    \label{fig:dif_lambda}
\end{figure*}

\begin{table}[t]
\centering
\caption{Quantitative comparison of our ablation methods. The ``Baseline'' method indicates the base inpainting model~\cite{fooocus_inpaint} combined with DreamBooth's finetuning strategy~\cite{dreambooth}. The ``*'' symbol indicates single-image training.}
\label{tab:ablation}
\fontsize{7}{8}

\renewcommand{\arraystretch}{1.2} 
\setlength{\tabcolsep}{3pt} 
\begin{tabular*}{\columnwidth}{ >{\centering\arraybackslash}cl>{\centering\arraybackslash}c >{\centering\arraybackslash}c>{\centering\arraybackslash}c }
\toprule
& \textbf{Method} & \textbf{CLIP-T↑} & \textbf{CLIP-I↑} & \textbf{DINO↑}  \\
\midrule
\multirow{6}{*}{\makecell{\textbf{ID} \\ \textbf{Pres.}}}
& Baseline   & 0.253 & 0.644 & 0.641 \\
& +DIF        &  0.291& 0.723& 0.735\\
& ++TAS       & 0.290& 0.728& 0.732 \\
\rowcolor[RGB]{235, 235, 235} & +++ADM & 0.285 & 0.728 & 0.730 \\

\cmidrule(lr){2-5} 
& DreamBooth*  & 0.248 & 0.609 & 0.614  \\
& \textbf{DreamMix*}    & 0.284 & 0.713 & 0.711 \\

\midrule
\multirow{6}{*}{\makecell{\textbf{Attr.} \\ \textbf{Editing}}}
& Baseline   & 0.235 & 0.601 & 0.612 \\
& +DIF        & 0.265 &  0.696 &  0.708 \\
& ++TAS       & 0.275 & 0.685 & 0.700  \\
\rowcolor[RGB]{235, 235, 235} & +++ADM & 0.289 & 0.674 & 0.695  \\
\cmidrule(lr){2-5}
& DreamBooth*  & 0.235 & 0.567 & 0.595 \\
& \textbf{DreamMix*}    & 0.284 & 0.659 & 0.672 \\
\bottomrule
\end{tabular*}

\end{table}

\textbf{Effect of $\lambda$ in DIF.} In DIF, we use $\lambda$ to control the separation of local content generation and global context harmonization. We conduct experiments with varying \(\lambda\) values and present the visual results in Figure~\ref{fig:dif_lambda}. As we can see, as \(\lambda\) increases, the inpainted regions become more harmonious with the context, but the allocation accuracy may decrease. Using only GCH (i.e., \(\lambda = 1\)) may pose generation failures (e.g., clock case) when inpainting on small sizes. Conversely, relying solely on local inpainting (i.e., \(\lambda = 0\)) can affect the overall harmony of the inpainted regions. 


\textbf{Additional Experiments on ADM and TAS Configurations.}
As shown in Table~\ref{tab:consolidated}, we conducted a series of additional experiments to evaluate the performance of our system under various ADM and TAS configurations. First, we replaced ChatGPT-4o with alternative VLMs in both ADM and TAS modules. Next, we evaluated the impact of varying the number of regularization images. Finally, we retrained the model using five different randomized attribute orders. Overall, the system showed stable performance across these configuration variations, demonstrating its robustness.

\begin{table*}[h]
\centering
\caption{Consolidated table showing performance across different VLMs, regularization image counts, and attribute orders.}
\label{tab:consolidated}
\resizebox{\textwidth}{!}{
\centering
\begin{tabular}{lcccc}
\toprule
\textbf{Category} & \textbf{Attri. Edit} & \textbf{CLIP-T} & \textbf{CLIP-I} & \textbf{DINO}  \\
\midrule
\multirow{4}{*}{VLM} 
  & \textbf{Qwen2-VL-2B~\cite{Qwen2VL, Qwen2VL2b}} & 0.287 & 0.675 & 0.695 \\
  & \textbf{LLaVA-v1.5-7B~\cite{llava_7b_v15}} & 0.290 & 0.674 & 0.692  \\
  & \textbf{SmolVLM-0.5B~\cite{smolvlm,smolvlm_05b}} & 0.287 & 0.672 & 0.694 \\
  & \cellcolor[gray]{0.9}\textbf{ChatGPT-4o~\cite{chatgpt}} & \cellcolor[gray]{0.9} 0.289 & \cellcolor[gray]{0.9} 0.674 & \cellcolor[gray]{0.9} 0.695  \\
\midrule
\multirow{4}{*}{Reg. Img.} 
  & \textbf{10} & 0.285 & 0.677 & 0.696 \\
  & \textbf{20} & 0.288 & 0.676 & 0.695  \\
  & \cellcolor[gray]{0.9}\textbf{30} & \cellcolor[gray]{0.9} 0.289 & \cellcolor[gray]{0.9} 0.674 & \cellcolor[gray]{0.9} 0.695 \\
  & \textbf{50} & 0.290 & 0.675 & 0.694 \\
\midrule
\multirow{1}{*}{Order} 
  & \textbf{random} $\Delta$ & ±0.002 & ±0.004 & ±0.002 \\
\bottomrule
\end{tabular}
}
\end{table*}

\begin{table}[t]
\centering
\caption{Impact of hyperparameter variations on model performance.}  
\label{tab:hyper_param_ablation}
\setlength{\tabcolsep}{8pt}
\begin{tabular}{ccccccc}
\toprule
\textbf{\(\textstyle \tau_1\)} & \textbf{\(\textstyle \tau_2\)} & \textbf{\(\textstyle \beta\)}
&\textbf{CLIP-T}&\textbf{CLIP-I}&\textbf{DINO} \\
\midrule
\cellcolor[gray]{0.9}1.5 & \cellcolor[gray]{0.9}0.7 & \cellcolor[gray]{0.9}0.4 & \cellcolor[gray]{0.9}0.289 & \cellcolor[gray]{0.9}0.674 & \cellcolor[gray]{0.9}0.695  \\
1.2 & 0.8 & 0.4 & 0.288 & 0.674 & 0.695 \\
1.0 & 1.0 & 0.4 & 0.286 & 0.675 & 0.696  \\
1.5 & 0.7 & 0.2 & 0.288 & 0.673 & 0.696 \\
1.5 & 0.7 & 1  & 0.285 & 0.672 & 0.693 \\
\bottomrule
\end{tabular}
\end{table}

\textbf{Hyper-parameter Ablation Studies.} 
The value selected for $\tau_1$, $\tau_2$ in Eq.~7 and $\beta$ in Eq.~8 are grounded on empirical observations. To further explore their influence, We altering these parameters empirically as detailed in Table~\ref{tab:hyper_param_ablation}. Our results demonstrate that while these adjustments had a negligible effect on the final performance metrics, they did impact the model's convergence speed. Specifically, smaller or larger values of the parameters slightly accelerated or slowed down the convergence, but the overall performance remained consistent across different settings. This suggests that the model is relatively robust to variations in these hyperparameters.

\textbf{Training on the Single Image.} To further validate the effectiveness of DreamMix with fewer training data, we retrain our DreamMix and DreamBooth with just one image per subject. As shown in Table~\ref{tab:ablation}, compared to the DreamBooth baseline, DreamMix indicates less performance degradation due to the incorporation of three key components. More importantly, even with only one image, DreamMix remains competitive with state-of-the-art approaches in Table~\ref{tab:all}. 

\subsection{Other Applications}

\begin{figure}[h]
    \includegraphics[width=0.48\textwidth]{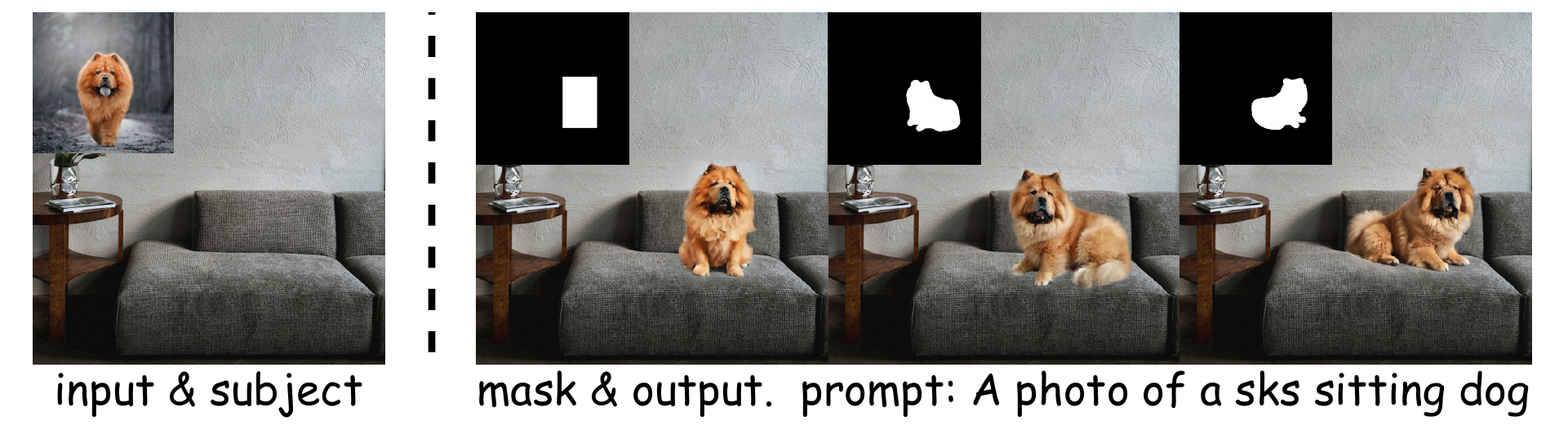}
    \caption{Examples of fine-grained pose control using masks. }
    \label{fig:pose_ctrl}
\end{figure}

\textbf{Pose Control Using Masks.} Fine-grained pose control can be achieved by leveraging specific masks for editing selected regions of an object. Unlike previous inpainting methods \cite{sdxl_inpaint, mimicbrush}, which are trained by randomly generating masks, we use SAM to obtain accurate boundary masks of objects and inflate them randomly, which not only greatly accelerates the fitting speed of the model but also gives the model the ability to use the shape of the mask to control the pose of the generated object as shown in Figure~\ref{fig:pose_ctrl}.

\begin{figure}[h]
    \includegraphics[width=0.48\textwidth]{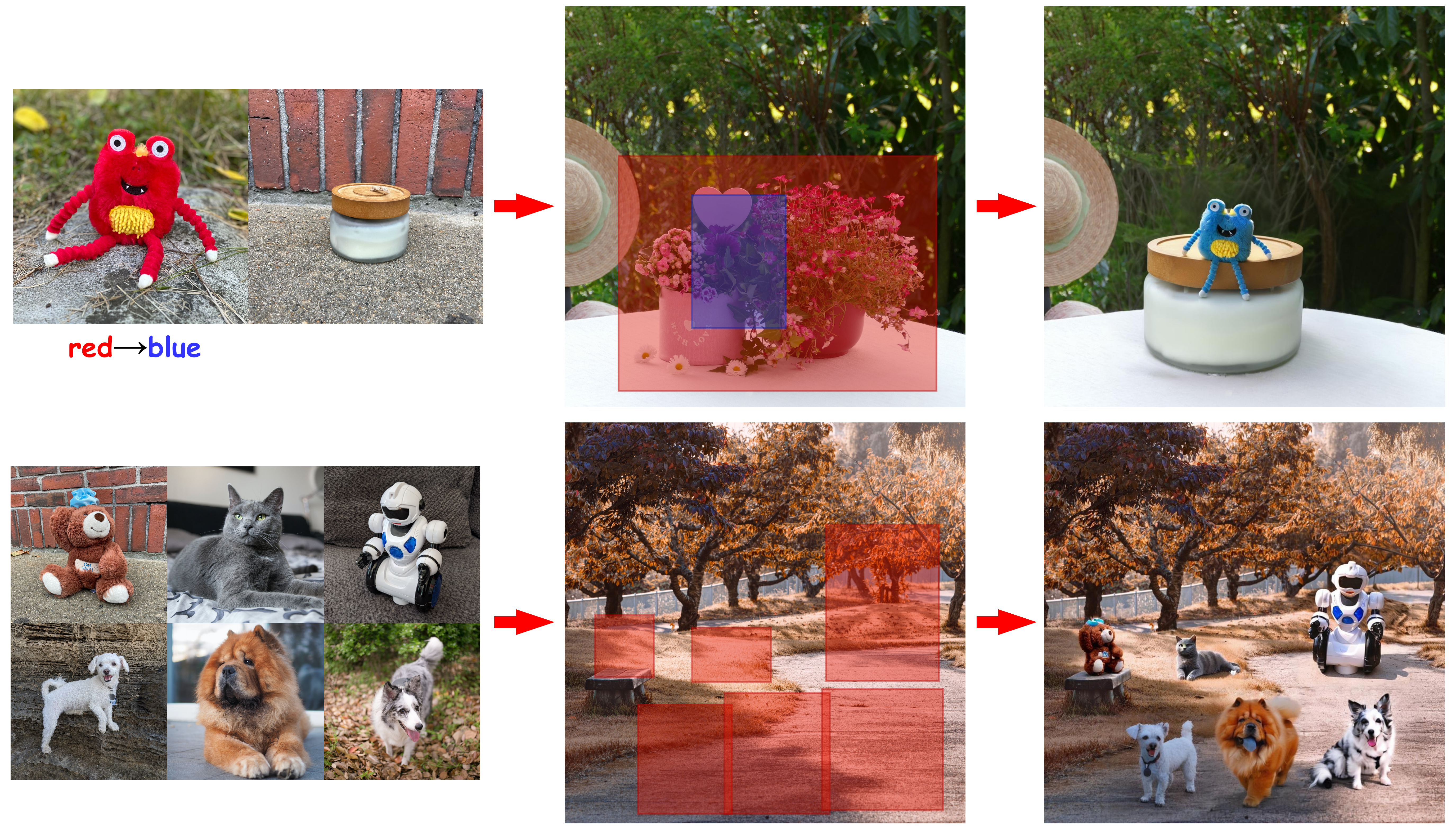}
    \caption{Visual examples of multi-subject inpainting, where multiple objects can be seamlessly integrated into a scene with interactive and editable attributes.}
    \label{fig:mutil_subject}
\end{figure}

\textbf{Multi-Subject Inpainting.}
Since our method cannot perform multi-subject inpainting in an end-to-end manner, we composite each subject on an image iteratively to support this task. As shown in Figure~\ref{fig:mutil_subject}, this approach allows our method to effectively composite multiple objects into a scene, facilitating object interaction and enabling detailed attribute editing.

\section{Conclusion}
\label{sec:conclusion}
We present a diffusion-based generative model for subject-driven image inpainting using both text and image guidance. An attribute decoupling mechanism and a textual attribution substitution module are proposed to facilitate text-driven editing of object attributes. Additionally, a disentangled inpainting framework seamlessly incorporates local subject allocation and global context harmonization to improve object insertion accuracy. Experiments demonstrate the superiority of the proposed method across various subject-driven inpainting applications. 
One limitation of our method is that relying solely on text and image guidance may pose challenges in achieving a harmonious interaction with the environment. Incorporating additional conditions (e.g., pose, depth) might solve this issue. Besides, extending our model to support multi-subject inpainting is a future direction.  

\section{Declarations}
\begin{itemize}
    \item The authors declare no competing interests.
    \item Data sharing is not applicable to this article as no new data were created or analyzed in this study.

\end{itemize}


\bibliography{sn-bibliography.bib}


\end{document}